\documentclass{article}

\usepackage{microtype}
\usepackage{graphicx}
\usepackage{mathtools}
\usepackage{amsmath,amssymb,amsthm}

\usepackage{booktabs} 

\usepackage{hyperref}
\usepackage{amsfonts}
\usepackage{amsmath,amssymb}

\DeclarePairedDelimiter{\norm}{\lVert}{\rVert}

\newcommand{\mbR}{\mathbb{R}}

\newcommand{\mcM}{\mathcal{M}}

\newcommand{\ulf}{\underline{f}}
\newcommand{\ulu}{\underline{u}}

\newcommand{\bp}{\mathbf{p}}
\newcommand{\bdel}{\mathbf{\delta}}

\usepackage{subcaption}
\usepackage{caption}

\usepackage{icml2019}

\title{Do Deep Minds Think Alike? Selective Adversarial Attacks for Fine-Grained Manipulation of Multiple Deep Neural Networks}

\author{Zain Khan, Jirong Yi, Raghu Mudumbai, Xiaodong Wu, Weiyu Xu\footnote{Z. Khan, J. Yi, R. Mudumbai, X. Wu, and W. Xu are with the Department of Electrical and Computer Engineering at University of Iowa, IA, USA, 52242. Corresponding emails: rmudumbai@engineering.uiowa.edu, xiaodong-wu@uiowa.edu, weiyu-xu@uiowa.edu.}}

\begin{document}
\maketitle

\begin{abstract}

 Recent works have demonstrated the existence of {\it adversarial examples} targeting a single machine learning system.  In this paper we ask a simple but fundamental question of ``selective fooling'': given {\it multiple} machine learning systems assigned to solve the same classification problem and taking the same input signal, is it possible to construct a perturbation to the input signal that manipulates the outputs of these {\it multiple} machine learning systems {\it simultaneously} in arbitrary pre-defined ways? For example, is it possible to selectively fool a set of ``enemy'' machine learning systems but does not fool the other ``friend'' machine learning systems?  The answer to this question depends on the extent to which these different machine learning systems ``think alike". We formulate the problem of ``selective fooling'' as a novel optimization problem, and report on a series of experiments on the MNIST dataset. Our preliminary findings from these experiments show that it is in fact very easy to selectively manipulate multiple MNIST classifiers simultaneously, even when the classifiers are identical in their architectures, training algorithms and training datasets except for random initialization during training. This suggests that two nominally equivalent machine learning systems do not in fact ``think alike" at all, and opens the possibility for many novel applications and deeper understandings of the working principles of deep neural networks.
\end{abstract}

\section{Introduction} \label{Sec:intro}
Recent works \cite{bhambri_study_2019,brunner_guessing_2018,carlini_towards_2017,kurakin_adversarial_2016,goodfellow_explaining_2014,szegedy_intriguing_2013} have shown that the outputs of many machine learning systems can be 
significantly changed using very small carefully-chosen perturbations to the inputs. In many 
cases, it is possible to gain essentially complete control over the outputs of such systems
with input perturbations that are imperceptible to human senses. 

Such vulnerability to {\it adversarial attacks} poses a major challenge to building robust
and reliable machine learning systems, and, as a result, understanding and mitigating this
phenomenon has become a major focus of research. While this research 
has produced many interesting findings \cite{tramer_space_2017,carlini_towards_2017,fawzi_adversarial_2018,brunner_guessing_2018,yi_trust_2019,bhambri_study_2019}, theoretical explanations for, and practical defenses against, adversarial attacks both remain open problems at this time.

If we imagine machine learning systems to have {\it perceptions}, adversarial attacks 
can be thought of as {\it illusions}. This gives rise to the question: to what
extent do different machine learning systems ``see" the same features? In particular, we may ask whether two machine learning systems share the same illusions. We seek to answer
this question by studying the extent to which it is possible to selectively manipulate multiple machine learning systems in different ways. In other words, we would like to know
if it is possible to design input perturbations that simultaneously manipulate multiple machine learning systems into ``seeing" different things. 

The study of such more powerful attacks may help us understand the nature of
adversarial attacks better. In addition, a study of such precise adversarial attacks may be of interest in its own right. For instance, the ability to precisely change the outputs of  a set of selected machine learning systems while leaving other systems unaffected may 
allow the creation of adversarial attacks that fool ``enemy'' machine learning systems, but spare ``friend'' machine learning systems. This can potentially be useful in various security applications, for example, in evading detections by enemy forces in battlefields.

To our knowledge, adversarial attacks that selectively and differentially target multiple 
machine learning systems have not been studied in detail in previous works. However, many of the underlying ideas have been used in other contexts. Certainly, the idea of
using multiple machine learning systems to augment each other in various ways in adversarial 
settings is not new. Many proposed defenses against adversarial attacks use auxiliary 
neural networks in various ways \cite{yin_divide-and-conquer_2019,yi_trust_2019,song_pixeldefend:_2018,samangouei_defense-gan:_2018}, e.g., binary classifiers that are trained to detect 
adversarial inputs, or generative networks for retraining networks to reduce their 
vulnerability to such inputs. A common way to train adversarial generative networks is by using
a discriminator network adversarially.

There also exists a significant volume of literature on {\it transferability} of adversarial attacks \cite{zheng_efficient_2019,lin_nesterov_2019,kang_transfer_2019,adam_reducing_2019,fawzi_adversarial_2018,tramer_space_2017,liu_delving_2016}, i.e., 
the question of whether an attack designed for one machine learning system is effective 
against another system. Our formulation in this paper is related to transferability but is 
very different and significantly more general. Specifically, we consider adversarial perturbations that are 
simultaneously designed to attack multiple machine learning systems, and also to have 
potentially {\it different} effects on each attacked system. Before our investigation, even the existence of such 
highly precise and selective attacks is not clear. Our main objective in this paper is to investigate this selective attack, and perform an experimental study to provide preliminary answers to this and other 
related questions.

\subsection{Summary of Objectives and Findings}

Our overall objective is to construct adversarial attacks which selectively and precisely manipulate multiple machine learning systems. In particular, we seek general answers to the following fundamental questions.

\begin{enumerate}

\item How selective and precise can adversarial attacks be? Specifically, what is the largest number of different machine learning systems that can be simultaneously and selectively manipulated by the same adversarial input?

\item A trivial mathematical observation is that selective adversarial attacks targeting  multiple machine learning systems simultaneously must require strictly larger perturbations than attacks targeting just one system. However, it is not clear how strong this relationship is. Thus, we would like to quantify the trade-off between how precise and selective an attack is, and
how imperceptible it is to human and other senses.

\item We would like to understand the transferability of selective adversarial perturbations i.e. how selective adversarial perturbations designed to target a number of machine learning algorithms affect other machine learning systems that the attack was not designed for.

\item Finally, we would like to use selective adversarial attacks as a tool to gain insights into the internal workings of otherwise-opaque machine learning algorithms.

\end{enumerate}

While our problem formulation is quite general, our experiments were focused on a set of $K=5$ deep neural network classifiers for the MNIST dataset\footnote{http://yann.lecun.com/exdb/mnist/}. These $5$ classifiers are identical in their architectures, training procedures and data sets, differing only in their random initial weights. Each of the $5$ classifiers also had approximately equal (and very high) accuracy for their test data sets. Our preliminary findings from these experiments are summarized as follows.

\begin{enumerate}

\item The answer to whether we can construct adversarial attacks selectively targeting multiple machine learning systems simultaneously is a definite 'YES'. We propose novel optimization formulation to design adversarial attacks which selectively attack multiple machine learning systems. 
We have been able to successfully generate adversarial perturbations against $5$ MNIST classifiers to several sets of pre-specified target labels. For example, we can design attacks which selectively fool a set of ``enemy'' machine learning systems, but do not fool the other ``friend'' machine learning systems.

\item Mathematically we can show that the size of the optimal perturbations increases with the number of classifiers being targeted, a conclusion that seems intuitively reasonable. However, our algorithm for generating these perturbations in some cases actually yields small perturbations when attacking more systems. We offer some possible explanations for this counter-intuitive result.

\item Adversarial examples that are designed to attack multiple classifiers have significantly greater {\it transferability} to other classifiers.

\end{enumerate}

The rest of the paper is organized as follows. Section \ref{Sec:back} presents a brief survey of related works. Selective adversarial attacks are formally defined as an optimization problem in Section \ref{Sec:prob}. Our experiments on selective adversarial attacks on $5$ MNIST classifiers are described in Section \ref{Sec:exp} . Section \ref{Sec:Conclusions} concludes this paper.

{\bf Notations}: We use $[N]$ to denote the set $\{1,2,\cdots,N\}$.

\section{Background and Related Work} \label{Sec:back}
As noted earlier, the phenomenon of adversarial vulnerability has now become an important focus of research in machine learning since the work by Szegedy et al. \cite{szegedy_intriguing_2013} and there is now a vast and growing literature on
this subject. The adversarial vulnerability has been observed in various tasks \cite{wang_backdoor_2020,zhou_fake_2019,bhambri_study_2019,zhao_blackbox_2019,li_adversarial_2019-1,khormali_copycat:_2019,carlini_audio_2018,grosse_adversarial_2016} and for different machine learning systems especially deep neural networks \cite{zugner_adversarial_2019,ko_popqorn:_2019}.   

Adversarial attacks are a broad phenomenon that can be classified in several possible ways. One important distinction is between white-box attacks where the attacker has access to the internals of the machine learning systems being targeted \cite{szegedy_intriguing_2013,goodfellow_explaining_2014,moosavi-dezfooli_deepfool:_2016,carlini_towards_2017,madry_towards_2017,bose_generalizable_2019} and corresponding defenses, \cite{meng_ensembles_2020,zhang_defending_2019,yi_trust_2019,shumailov_sitatapatra:_2019,samangouei_defense-gan:_2018,xie_feature_2018,madry_towards_2017,ilyas_robust_2017}, and black box attacks \cite{yan_subspace_2019,meunier_yet_2019,brunner_guessing_2018,papernot_practical_2017,liu_delving_2016} and their defenses \cite{shumailov_sitatapatra:_2019,xie_feature_2018,ilyas_robust_2017}.

The fast gradient sign method (FGSM) \cite{szegedy_intriguing_2013,goodfellow_explaining_2014,kurakin_adversarial_2016,yi_trust_2019}, the DeepFool method \cite{moosavi-dezfooli_deepfool:_2016}, and the Carlini-Wagner (CW) attack \cite{carlini_towards_2017} are three of the most popular algorithms for generating white-box attacks.

In adversarial attacks based on FGSM \cite{szegedy_intriguing_2013,goodfellow_explaining_2014}, an adversarial sample is generated via
\begin{align}\label{Defn:FGSM}
x_{adv} = {\rm trunc}\left(x + \epsilon \cdot{\rm sign}(\nabla_x \ell(x,u))\right),
\end{align}  
where $x\in\mbR^M$ is a benign sample, $\epsilon>0$ is a small positive constant, ${\rm sign}(\cdot)$ is an element-wise sign function, and $\nabla_x \ell(x,u)$ is the gradient of loss function $\ell$ with respect to the input $x$. The $u\in\{1,2,\cdots,L\}$ is the true label of $x$, and the ${\rm trunc}(\cdot)$ function is applied to guarantee that the adversarial sample is within a normal range, i.e., the pixels of an image should be within $[0,255]$.

It has been observed that adversarial samples designed for one machine learning system can also fool other machine learning systems \cite{liu_delving_2016,papernot_practical_2017,tramer_space_2017,dong_evading_2019}. This transferability of adversarial samples allows us to design adversarial samples for a substitute system, and then use these samples to fool the target system. In sampling-based black-box attacking method, we initialize an adversarial sample which can be quite different from the benign sample, and then we gradually shrink the distance from the adversarial sample to the benign sample by randomly sampling a moving direction \cite{brunner_guessing_2018,narodytska_simple_2016}. Different from the white-box attacks where the gradient can be easily calculated, the gradient-free black-box attack uses an estimate of the gradient to perform the searching of adversarial samples \cite{chen_zoo_2017,ilyas_black-box_2018,tu_autozoom_2019}.

As one of the most important black-box attacking methods, the transferability of adversarial samples has received tremendous amount of attention recently \cite{shumailov_sitatapatra:_2019,zheng_efficient_2019,lin_nesterov_2019,kang_transfer_2019,adam_reducing_2019,fawzi_adversarial_2018,tramer_space_2017,liu_delving_2016}. In \cite{liu_delving_2016}, Liu et al. reported that though an adversarial sample generated for one system can fool another system, it cannot always fool the two systems into making the same wrong decision, e.g., being classified as in the same wrong class. This means that the adversarial samples can transfer, but their labels cannot always. The authors further proposed an ensemble-based approach to increase the transferability of targeted adversarial attacks to a single to-be-attacked black-box machine learning system.  This topic was further studied by Adam et al. \cite{adam_reducing_2019}. In \cite{adam_reducing_2019}, the authors proposed to optimize the distance of the gradients of different models in training so that the transferability of adversarial attacks can be reduced. Another viewpoint of the adversarial transferability was proposed by Lin et al. \cite{lin_nesterov_2019}, and they argued that the transferability of an adversarial sample is just like the generalization performance of a training learning system. Based on this, they proposed to improving the transferability via techniques similar to those used for improving the generalization performance of learning systems. 

On the theoretical side, Tramer et al. made attempts to investigate the fundamental reasons accounting for the tranferability of adversarial samples \cite{tramer_space_2017}. They showed that for two different models, their decisions regions are highly overlapped, and also gave conditions on the data distribution for guaranteeing the transferability. Another line of theoretical work by Fawzi et al. \cite{fawzi_adversarial_2018} gave an upper bound on the probability for an adversarial sample to be transferable when the decision regions of learning systems are highly overlapped. 

This paper focuses on white-box evasion attacks on classifier networks where the attacks can be optimally designed using iterative algorithms. These are, of course, the most favorable conditions for the attacker and allows for the design of the strongest and most powerful class of attacks.

\section{Problem Statement: Optimal Selective Adversarial Attacks} \label{Sec:prob}

We now formally define selective adversarial attacks as a constrained optimization problem.

Consider a set $\mcM \doteq \{M_1,~M_2,~\dots,~M_N \}$ of $N$ trained machine learning classifiers. The $i$-th ($i=1 \dots N$) machine learning system $M_i$ maps an input vector $x \in \mbR^M$ to one label $u$: $f_i(x)=u \in \{1 \dots L\}$, where $L$ is the number of labels, and $\{1 \dots L\}$ is the set of labels. Let $f_0(): \mbR^M \rightarrow \{1 \dots L\}$ denote the ground-truth mapping, namely $u_0=f_0(x)$ is the ``true label" of a given input $x$.

\begin{figure}
\centering
\includegraphics[width=0.45\textwidth]{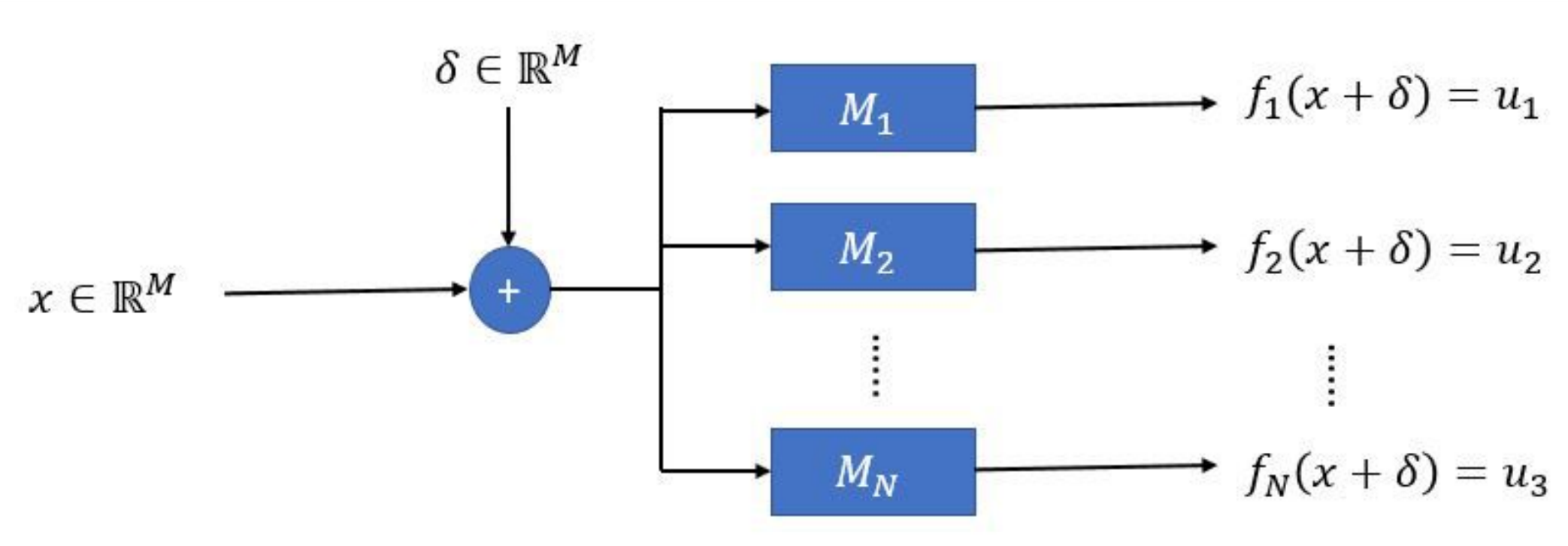}
\caption{Conceptual framework of selective adversarial attack.}\label{Fig:SelectiveAdversarialAttack}
\includegraphics[width=0.45\textwidth]{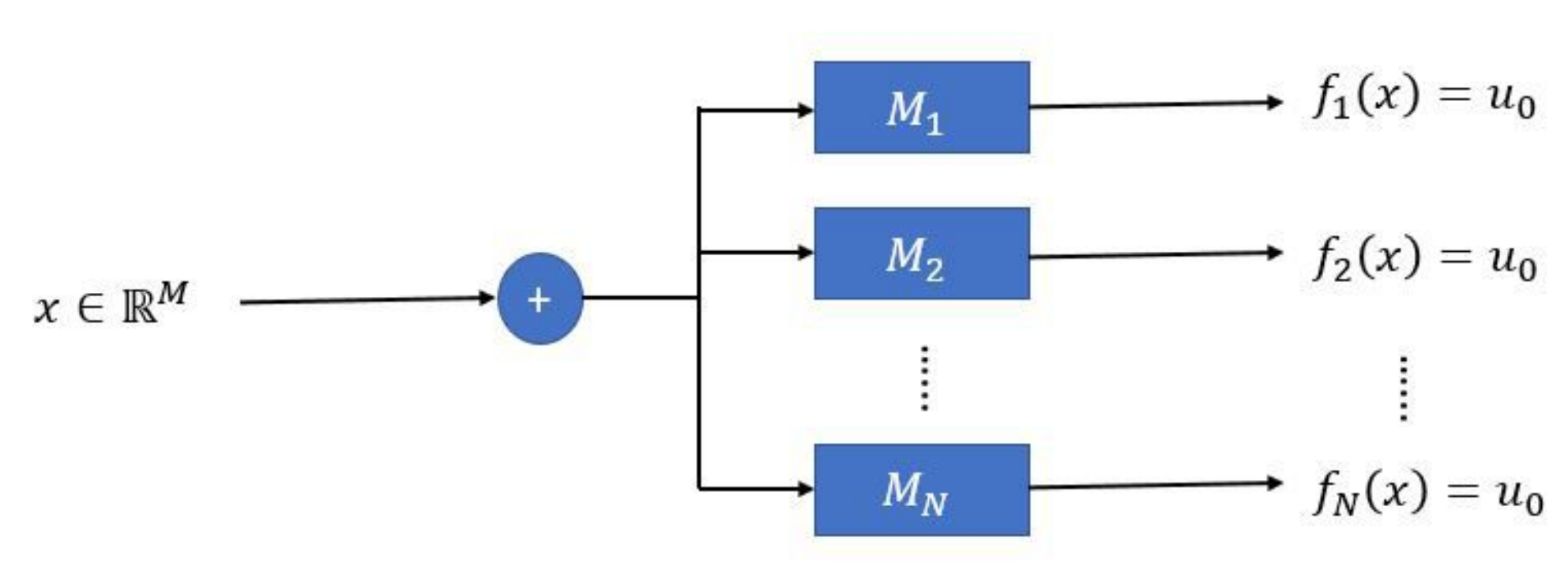}
\caption{Conceptual framework of classification with multiple classifiers.}\label{Fig:ClassificationWithMultipleClassifiers}
\end{figure}

Machine learning classifiers typically work with high-dimensional inputs i.e. $M \gg 1$. We assume that each of the classifiers $\mcM_i$ are highly accurate i.e.:
\begin{align}
\Pr \left( f_i(x) = f_0(x) \right) &\approx 1,~\forall i=1 \dots N
\end{align}
over a relevant test distribution $p(x)$ of the inputs $x$. Let $\ulf(\cdot)$ denote the vector function mapping an input $x \in \mbR^M$ to a vector of $N$ labels $\ulu =[u_1~u_2~\dots u_N] \in \{1 \dots L\}^N$ i.e. $\ulu = \ulf(x) \doteq [f_1(x)~f_2(x)~\dots~f_N(x)]$. See Figure \ref{Fig:SelectiveAdversarialAttack} and \ref{Fig:ClassificationWithMultipleClassifiers}.

Consider some arbitrary input vector $x_0 \in \mbR^M$, the corresponding true label $u_0 \doteq f_0(x_0)$, and an arbitrary set of $N$ ``target labels" $\ulu_t \in \{1 \dots L \}^N$. Note that if $x_0$ is chosen from the test distribution $p(x)$, then with high probability, all the $N$ classifiers produce the correct label to the unperturbed input vector $x_0$ i.e.
\begin{align}
\Pr \left( \ulf(x_0) =[u_0~u_0~\dots~u_0] \right) &\approx 1
\end{align}
We now formally define an optimal selective adversarial attack as the smallest perturbation applied to the input vector $x_0$ that causes the $N$ classifiers $\mcM_i,~i=1 \dots N$ to output the target labels $\ulu_t$, each of which may be different from (or remains the same as) the true label $u_0$, and may be different from each other. Specifically the optimal selective perturbation $d(x_0, \ulu_t): \mbR^M \times \{1 \dots L \}^N \rightarrow \mbR^M$ for an input vector $x_0$ and the intended label set $\ulu_t$ is defined as:
\begin{align}
d(x_0, \ulu_t) &\doteq \arg \min_{\delta \in \mbR^M} \norm{\delta},~\nonumber \\
\mathrm{subject~to~} & \ulf \left( x_0 + \delta \right) \equiv \ulu_t \label{eq:opt1}
\end{align}

{\bf Remark.} Note that in the definition (\ref{eq:opt1}), there is an implicit assumption that the constraint is feasible i.e. that there exists $x \in \mbR^N$ such that $\ulf(x) \equiv \ulu_t$ for any target label set $\ulu_t$. This, of course, is an assumption about the range of the functions $\ulf()$ describing the classifiers. Indeed it is a very strong assumption about how differently multiple nominally equivalent classifiers process their inputs. One of our objectives in this paper is to empirically test this assumption. Our findings in this paper surprisingly show that the constraint is often feasible under an arbitrary set of labels.

\subsection{Selective Fooling} \label{sec:selfool}

We now consider an interesting sub-class of selective adversarial attacks which we will call
Selective Fooling. In this class of attacks, we choose a single adversarial target label $u_a$ that
is different from the true label $u_0$ and design a perturbation that will cause a subset of 
$n < N$ classifiers in $\mcM$ to output the label $u_a$ while the remaining $N-n$ classifiers
in $\mcM$ continue to output the true label $u_0$. In other words, we seek to {\it selectively
fool} a subset of classifiers to output a particular target label $u_a$ while leaving the remaining 
ones unaffected.

Formally, Selective Fooling is a selective adversarial attack where the target label vector is
set to be:
\begin{align}
\ulu_t[i]&=
\begin{cases}
&u_a,~i \in \{1 \dots n\}, \\
&u_0~i \in \{n+1 \dots N\}
\end{cases} \label{eq:selfool}
\end{align}
for some $u_a \in \{1 \dots L\} \setminus  \{u_0\}$.

\subsection{Selective Adversarial Attacks using Gradient Descent}

We now discuss a simple method that extends previously known algorithms such as FGSM \cite{szegedy_intriguing_2013} that have been shown to be effective in generating adversarial examples to create selective adversarial perturbations. The basic idea is to formulate a cost function that serves as a proxy for the optimization problem in (\ref{eq:opt1}), and then use a gradient descent procedure to approximate a solution to (\ref{eq:opt1}).

Let $\bp_i(x),~i=1 \dots N,~x \in \mbR^M$ represent the probability distribution output by the classifier $M_i$ for the input vector $x$ over the $L$ possible labels $1 \dots L$. Mathematically $\bp_i(x)$ is a $1 \times L$ vector that satisfies $\sum_{u=1}^L \bp_{i,u}(x) \equiv 1,~\forall i, x$, where $\bp_{i,u}(x)$ is the $u$-th element of $\bp_i(x)$. In practice, $\bp_i$ are typically calculated using a ``softmax" function \cite{goodfellow_deep_2016} in the last layer of a deep neural network. Let $\bdel_u,~u \in \{1 \dots L\},$ be the $1 \times L$ Kronecker delta vector i.e.:
\begin{align}
\bdel_u[l] &\doteq 
\begin{cases}
1,&~l=u, \\
0,~&\mathrm{otherwise}
\end{cases}
\end{align}

Given a target label set $\ulu=[u_1~u_2~\dots~u_N]$, we define the following cost function $J(x),~x \in \mbR^M$:
\begin{align}
J(x) &= \sum_{i=1}^N \norm{\bp_i(x) - \bdel_{u_i}}^2 \label{eq:obj}
\end{align}

Our selective adversarial attack algorithm will iteratively reduce this proxy cost function by gradient descent. Of course a method for generating adversarial examples via gradient searches on the heuristic cost function in (\ref{eq:obj}) can only approximate the optimal solution to (\ref{eq:opt1}). Indeed, we will see in our experimental results, that adversarial perturbations obtained from this cost function differs from the optimal solution in some interesting and counter-intuitive ways. We also remark that we can replace the cost function in (\ref{eq:obj}) with other cost functions, such as cost functions using cross-entropy.


\section{Experimental Results} \label{Sec:exp}

We constructed a total of $N=5$ Convolutional Neural Networks with the architecture shown in Table \ref{Tab:NNArchitecture}. We trained each of them on the MNIST handwritten digit recognition dataset\footnote{http://yann.lecun.com/exdb/mnist/}. All $N$ networks were identical except for a random initialization of their weights.

\begin{table}
    \centering
    \begin{tabular}{|c|c|}
    \hline
        Layer name & Layer type \\
        \hline
        Input layer & input(28,28,1)\\
        \hline
        Layer 1 & conv2d(32,3,3) + ReLU\\
        \hline 
        Layer 2 & conv2d(32,3,3) + ReLU + maxpool(2,2)\\
        \hline 
        Layer 3 & conv2d(64,3,3) + ReLU \\
        \hline
        Layer 4 & conv2d(64,3,3) + ReLU + maxpool(2,2)\\
        \hline 
        Layer 5 & FC(200) + ReLU\\
        \hline
        Layer 6 & FC(200) + ReLU\\
        \hline
        Layer 7 & FC(10) + Softmax\\
        \hline
    \end{tabular}
    \caption{Architecture of neural networks in experiments. The FC is the fully connected layer, and the stride is 1 for convolution and maxpooling.}
    \label{Tab:NNArchitecture}
\end{table}

\subsection{Training the Classifiers}

We used a Stochastic Gradient Descent to minimize the softmax cross entropy function with a learning rate of $0.01$, a decay of $1e-6$, and a momentum of $0.9$ to train each of the $5$ networks. The models were trained using the MNIST training data, obtained from the Keras MNIST dataset, for $10$ epochs with $55,000$ images per epoch. The trained networks were evaluated on $10,000$ MNIST images from a separate test set. This resulted in all of the $5$ networks achieving an accuracy of at least $95\%$ accuracy on the test set. The models were trained using different random initializations of the network weight and bias parameters. 

Next we describe the procedure we used to generate selective adversarial examples and a series of experiments that we performed using these trained networks.

\subsection{Constructing Selective Adversarial Attacks}

In all of our experiments, we used images with the true class label of `0' (results for other digits are similar, and so we focus on label `0' for our presentation ) from the aforementioned MNIST handwritten digit recognition dataset, and then constructed selective adversarial attacks designed to manipulate each of the $5$ networks in arbitrary pre-specified ways. We used modified versions of two popular algorithms to generate our adversarial examples: (a) the modified Carlini-Wagner (mCW) algorithm \cite{carlini_towards_2017} and (b) the modified Fast Gradient Signed Method (mFGSM) \cite{szegedy_intriguing_2013,kurakin_adversarial_2016}.

The original Carlini-Wagner algorithm\footnote{https://github.com/carlini/nn\_robust\_attacks} was designed to construct adversarial examples for a single classifier using a gradient descent procedure. We modified this algorithm to perform gradient descent on the loss function $J(x)$ as defined in (\ref{eq:obj}). The following parameter values were used in our modified CW algorithm: confidence value $0.1$, learning rate $0.02$, the number of binary search steps $9$, and the max number of iterations $10,000$. Our modified FGSM algorithm\footnote{https://github.com/soumyac1999/FGSM-Keras} also used the loss function very similar to (\ref{eq:obj}) (with the cross-entropy function replacing the $l_2$ norm in (\ref{eq:obj})). In our experiments with the modified FGSM algorithm, we set the parameter $\epsilon = 0.075$ with the maximum number of iterations fixed at $30$.

\begin{figure}
    \begin{center}
    \includegraphics[width=0.5\textwidth,height=4cm]{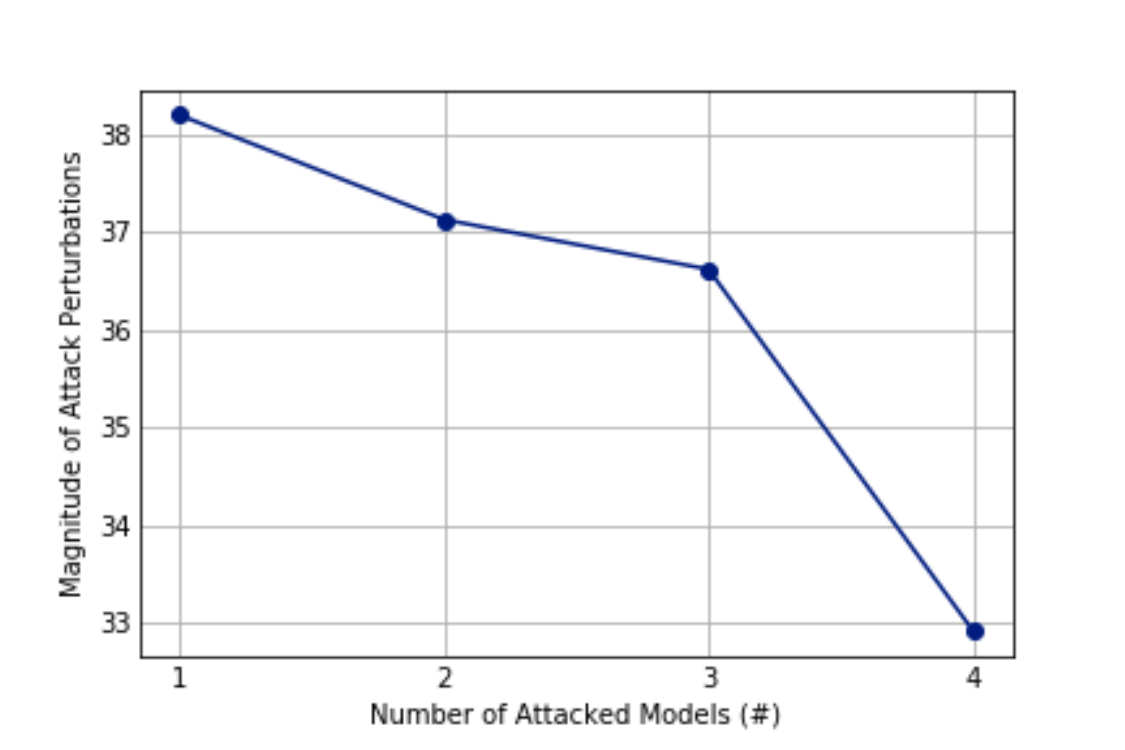}
    \caption{Average size of the adversarial perturbations generated by the mCW algorithm as a function of the number of attacked classifiers.}
    \label{fig:magnitudeCW4}
    \includegraphics[width=0.5\textwidth,height=4cm]{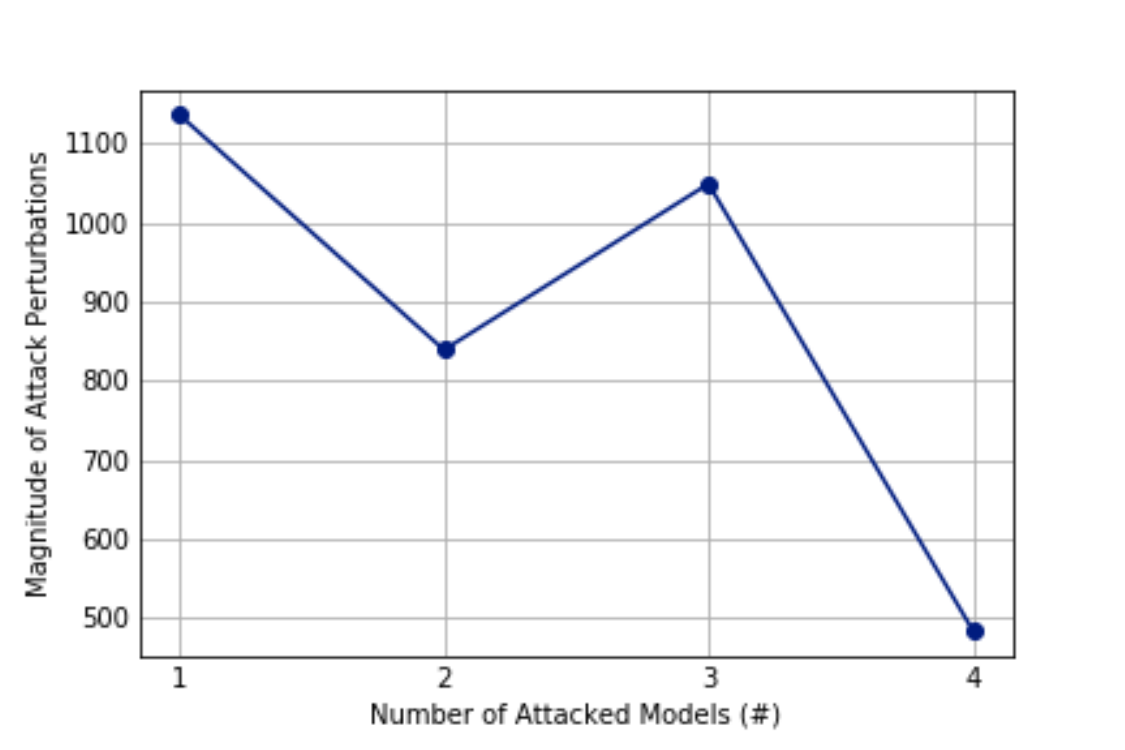}
    \caption{Average size of the mFGSM attack as a function of the number of attacked classifiers.}
    \label{fig:magnitudeFGSM4}
    \end{center}
\end{figure}

\subsection{Selective Fooling Experiment with Two Classifiers}

Our first set of experiments involved the simple sub-class of selective adversarial attacks defined in Section \ref{sec:selfool} with $N=2,~n=1$ i.e. we attempt to construct adversarial examples that successfully fool one of two classifiers while leaving the other classifier unaffected.

A total of $25$ images of the label $0$ were perturbed using both the mCW and mFGSM algorithms described earlier with random target labels. The mFGSM algorithm was able to successfully construct Selective Fooling attacks on all of the $25$ images. The mCW algorithm was also able to generate adversarial examples for a majority of images, but takes longer to converge than mFGSM.

This shows the {\it existence} of {\it selective} adversarial examples, at least for this simple special case of Selective Fooling attack, and suggests that it may indeed be feasible to construct adversarial examples that affect multiple classifiers in different ways (and surprisingly often in all the possible ways) and motivates a deeper study of selective adversarial attacks.


We also tested the effect of the adversarial examples generated by the mCW algorithm on a third classifier that was not involved in the construction of the attack; the confidence value of this third classifier in the true label ($0$) averaged over the $25$ images was calculated to be $0.778$. Roughly speaking, this shows that on average, a third classifier is approximately $75\%$ unaffected by the Selective Fooling attack. In contrast, the corresponding number for a non-selective fooling attack against just one classifier using the unmodified Carlini-Wagner algorithm was $0.843$ i.e. a third classifier was approximately $85\%$ unaffected by such a non-selective fooling attack. This seems somewhat counter-intuitive: we may expect that a selective attack that narrowly targets one classifier while leaving another classifier unaffected will be {\it less} likely to affect a third classifier. Below we consider a possible explanation for this and certain other similarly counter-intuitive results.

\begin{figure}[!htbp]
    \begin{center}
    \includegraphics[width=0.5\textwidth,height=4cm]{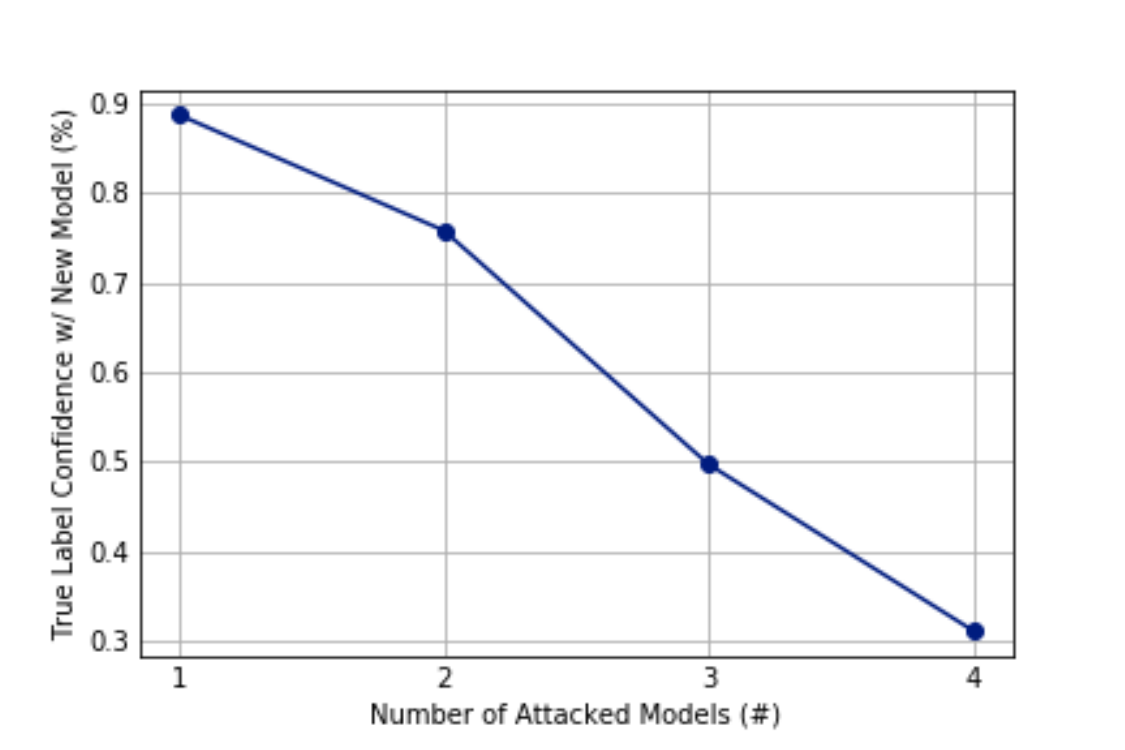}
    \caption{Confidence value in the true label for a fifth classifier as a function of the number of attacked classifiers.}
    \label{fig:transferabilityCW4}
    \includegraphics[width=0.5\textwidth,height=4cm]{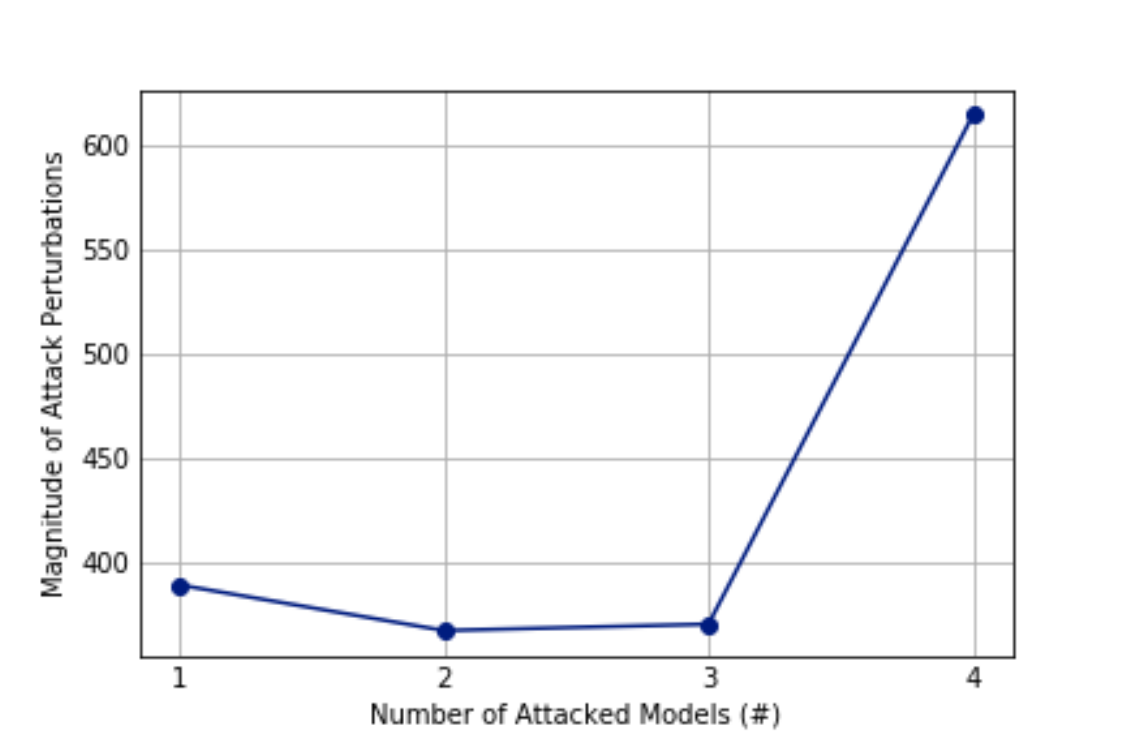}
    \caption{Magnitude of the FGSM attack as the number of attacked models increases and the number of defended model decreases when observing the general $M$ and $n$ attacks for four total models.}
    \label{fig:magnitudeFGSM4Precise}
    \end{center}
\end{figure}

\subsection{Selective Fooling Experiment with Four Classifiers}
\begin{figure}[!htbp]
    \begin{center}
    \begin{enumerate}
    \item[a)] \text{One Attacked, Three Defended}\\
    \vspace{1mm}
    \includegraphics[width=0.9\columnwidth,height=3cm]{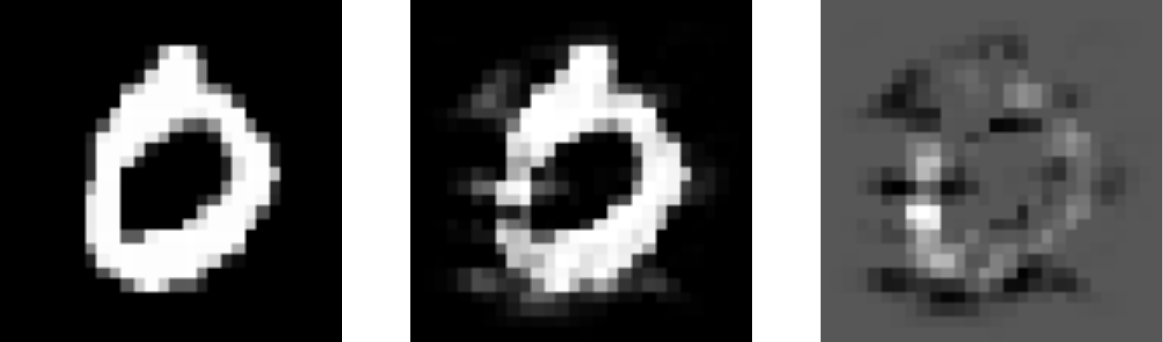}
    \item[b)] \text{Two Attacked, Two Defended}\\
    \vspace{1mm}
    \includegraphics[width=0.9\columnwidth,height=3cm]{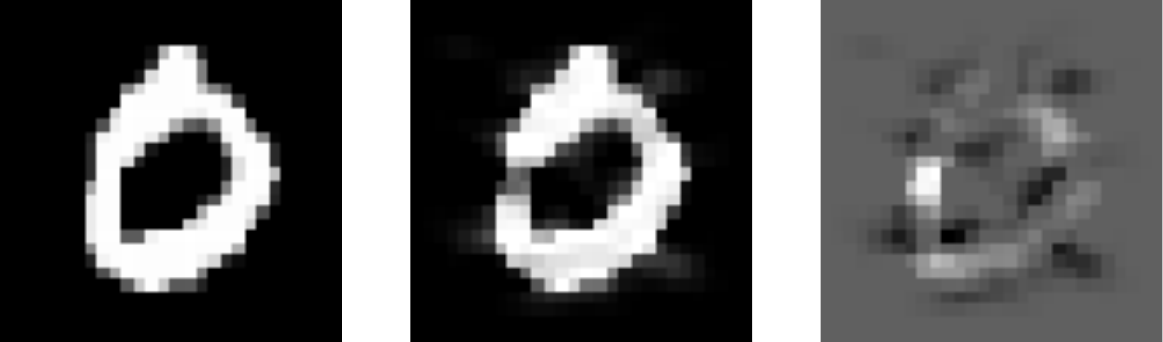}
    \item[c)] \text{Three Attacked, One Defended}\\
    \vspace{1mm}
    \includegraphics[width=0.9\columnwidth,height=3cm]{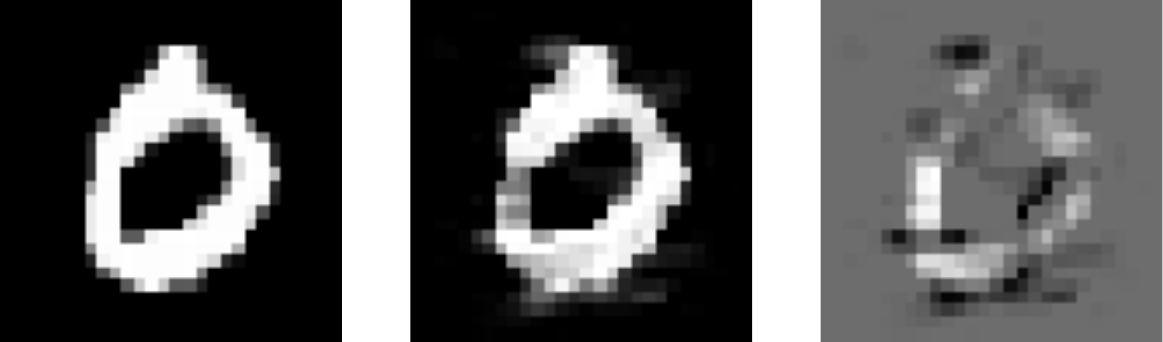}
    \item[d)] \text{Four Attacked, None Defended}\\
    \vspace{1mm}
    \includegraphics[width=0.9\columnwidth,height=3cm]{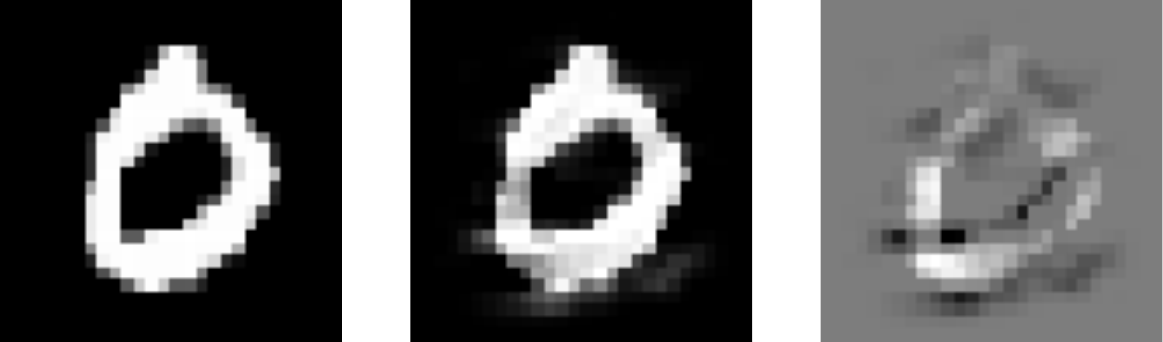} 
    \end{enumerate}
    \caption{Examples of input images, their adversarial counterpart, and the perturbations added for cases where a) $n$=1, b) $n$=2, c) $n$=3, and d) $n$=4 classifiers are attacked via the mCW algorithm out of $N=4$ total.}
        \label{fig:examples}
    \end{center}
\end{figure}

In our next set of experiments, we constructed a series of Selective Fooling attacks with $N=4$ classifiers. Specifically, for a set of $25$ images with the true label of $u_0=0$ and for each of the $15$ non-empty subsets of the $4$ classifiers, we attempted to construct selective adversarial attacks to drive their outputs to a target label of $u_a=9$ using both the mCW and mFGSM algorithms. Surprisingly, both the mCW and mFGSM methods were able to successfully construct adversarial examples that successfully fooled {\it each} of the $2^4-1=15$ non-empty subsets of the $N=4$ classifiers while keeping the complementary subset of classifiers unfooled, for each of the $25$ images.

\begin{figure*}[!htb]
    \begin{center}
    \includegraphics[width=\textwidth]{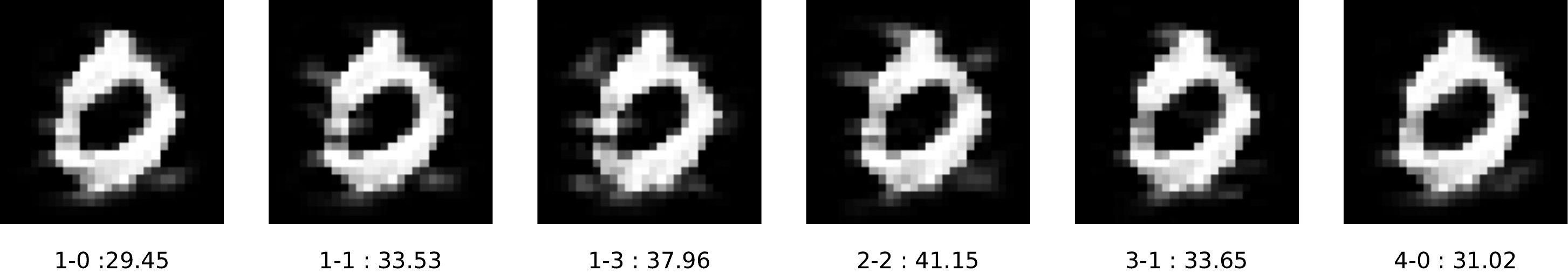}
    \caption{For an adversarial image attacking $m=N-n$ classifiers and defending $n$ classifiers, designated as $m-n$, we can see the magnitudes of the attacks for the unmodified Carlini-Wagner attack (1-0), the simplest attack and defend case (1-1), and the four subset cases (where the target labels are all set to $9$).}
    \label{fig:example_perturbations}
    \end{center}
\end{figure*}

The average size of the perturbations measured as the sum of the absolute value of pixel differences between the input image and the adversarial image, is shown as a function of the number of targeted classifiers $n$ in Figs. \ref{fig:magnitudeCW4} and \ref{fig:magnitudeFGSM4} for the mCW and mFGSM algorithms respectively.

{\bf Discussion.} The mere fact that there exists small perturbations which selectively and precisely attack every possible subset of classifiers is quite surprising. This suggests that these classifiers might look at very different features for classifications, and do not ``think'' alike, even though they share the same architectures and training data. Figures \ref{fig:magnitudeCW4} and \ref{fig:magnitudeFGSM4} suggest that on average we need {\it smaller} perturbations to attack more classifiers. We might expect that it would take larger perturbations to simultaneously change the outputs of two classifiers compared to one classifier.  One possible explanation for this observation - as well as our earlier observation about transferability of the Selective Fooling attack - is that the attacks generated by the mCW and mFGSM algorithms are not close to being optimal; a search constrained to leave one or more classifiers unaffected leads to smaller perturbation perhaps because it is forced to keep more features of the original image intact as compared to an unconstrained search.

The confidence in the true label from a fifth classifier that was not involved in the generation of the adversarial examples is shown as a function of the size of the target set $n$ in Figure \ref{fig:transferabilityCW4} for the mCW generated attacks. This plot is consistent with our intuitive expectation that an adversarial example that successfully attacks many classifiers simultaneously is also more statistically likely to transfer to a random classifier.

Adversarial example images for different values of $n$ are shown in Figure \ref{fig:examples}. Typical examples of adversarial images with varying $N$ and $n$ and their respective magnitudes can be seen in Fig. \ref{fig:example_perturbations}.

\subsection{Attacking and Defending Subsets of Four Classifiers with Different Target Labels}

Our final set of experiments involved a generalization of the Selected Fooling attack where a subset $n$ of $N=4$ classifiers were targeted to be attacked to $n$ different target labels chosen at random, while keeping the outputs of the remaining $N-n$ classifiers unchanged. Once again, the mFSGM algorithm was able to successfully generate adversarial examples for all $15$ non-empty subsets of the $N=4$ classifiers. The average size of the resulting perturbations as a function of $n$ is shown in Fig. \ref{fig:magnitudeFGSM4Precise}.

\section{Conclusions}\label{Sec:Conclusions}

We formulated the problem of selective adversarial attacks, and presented an experimental investigation of a generalized class of adversarial attacks that are designed to manipulate the outputs of multiple machine learning classifiers {\it simultaneously} in arbitrarily pre-defined ways. We formulated a novel optimization problem to search for such selective attacks, and applied modified versions of popular algorithms used for the construction of white-box adversarial examples to design targeted attacks on $5$ different trained classifiers for the MNIST dataset. Our results show that it is indeed possible to construct precisely targeted adversarial attacks that can arbitrarily modify the outputs of multiple classifiers while keeping another set of classifiers unaffected. These results motivate a deeper study of such selective adversarial attacks and also suggest that different classifiers that have nominally similar performance may in fact have very different decision regions. 




\bibliographystyle{icml2019} 
\bibliography{Ref_TSF}

\begin{thebibliography}{42}
\providecommand{\natexlab}[1]{#1}
\providecommand{\url}[1]{\texttt{#1}}
\expandafter\ifx\csname urlstyle\endcsname\relax
  \providecommand{\doi}[1]{doi: #1}\else
  \providecommand{\doi}{doi: \begingroup \urlstyle{rm}\Url}\fi

\bibitem[Adam et~al.(2019)Adam, Smirnov, Haibe-Kains, and
  Goldenberg]{adam_reducing_2019}
Adam, G., Smirnov, P., Haibe-Kains, B., and Goldenberg, A.
\newblock Reducing adversarial example transferability using gradient
  regularization.
\newblock \emph{arXiv:1904.07980 [cs.LG]}, April 2019.

\bibitem[Bhambri et~al.(2019)Bhambri, Muku, Tulasi, and
  Buduru]{bhambri_study_2019}
Bhambri, S., Muku, S., Tulasi, A., and Buduru, A.
\newblock A study of black box adversarial attacks in computer cision.
\newblock \emph{arXiv:1912.01667 [cs, stat]}, December 2019.
\newblock arXiv: 1912.01667.

\bibitem[Bose et~al.(2019)Bose, Cianflone, and
  Hamiltion]{bose_generalizable_2019}
Bose, A., Cianflone, A., and Hamiltion, W.
\newblock Generalizable adversarial attacks using generative models.
\newblock \emph{arXiv:1905.10864 [cs, stat]}, May 2019.
\newblock arXiv: 1905.10864.

\bibitem[Brunner et~al.(2018)Brunner, Diehl, Le, and
  Knoll]{brunner_guessing_2018}
Brunner, T., Diehl, F., Le, M., and Knoll, A.
\newblock Guessing smart: biased sampling for efficient black-box adversarial
  attacks.
\newblock \emph{arXiv:1812.09803 [cs, stat]}, December 2018.
\newblock arXiv: 1812.09803.

\bibitem[Carlini \& Wagner(2017)Carlini and Wagner]{carlini_towards_2017}
Carlini, N. and Wagner, D.
\newblock Towards evaluating the robustness of neural networks.
\newblock In \emph{2017 {IEEE} {Symposium} on {Security} and {Privacy} ({SP})},
  pp.\  39--57, May 2017.
\newblock \doi{10.1109/SP.2017.49}.

\bibitem[Carlini \& Wagner(2018)Carlini and Wagner]{carlini_audio_2018}
Carlini, N. and Wagner, D.
\newblock Audio adversarial examples: targeted attacks on speech-to-text.
\newblock \emph{arXiv:1801.01944 [cs]}, January 2018.
\newblock arXiv: 1801.01944.

\bibitem[Chen et~al.(2017)Chen, Zhang, Sharma, Yi, and Hsieh]{chen_zoo_2017}
Chen, P., Zhang, H., Sharma, Y., Yi, J., and Hsieh, C.
\newblock {ZOO}: zeroth order optimization based black-box attacks to deep
  neural networks without training substitute models.
\newblock \emph{Proceedings of the 10th ACM Workshop on Artificial Intelligence
  and Security - AISec '17}, pp.\  15--26, 2017.
\newblock \doi{10.1145/3128572.3140448}.
\newblock arXiv: 1708.03999.

\bibitem[Dong et~al.(2019)Dong, Pang, Su, and Zhu]{dong_evading_2019}
Dong, Y., Pang, T., Su, H., and Zhu, J.
\newblock Evading defenses to transferable adversarial examples by
  translation-invariant attacks.
\newblock \emph{arXiv:1904.02884 [cs.CV]}, April 2019.

\bibitem[Fawzi et~al.(2018)Fawzi, Fawzi, and Fawzi]{fawzi_adversarial_2018}
Fawzi, A., Fawzi, H., and Fawzi, O.
\newblock Adversarial vulnerability for any classifier.
\newblock \emph{arXiv:1802.08686 [cs, stat]}, February 2018.
\newblock arXiv: 1802.08686.

\bibitem[Goodfellow et~al.(2014)Goodfellow, Shlens, and
  Szegedy]{goodfellow_explaining_2014}
Goodfellow, I., Shlens, J., and Szegedy, C.
\newblock Explaining and harnessing adversarial examples.
\newblock \emph{arXiv:1412.6572 [cs, stat]}, December 2014.
\newblock arXiv: 1412.6572.

\bibitem[Goodfellow et~al.(2016)Goodfellow, Bengio, Courville, and
  Bengio]{goodfellow_deep_2016}
Goodfellow, I., Bengio, Y., Courville, A., and Bengio, Y.
\newblock \emph{Deep learning}, volume~1.
\newblock MIT press Cambridge, 2016.

\bibitem[Grosse et~al.(2016)Grosse, Papernot, Manoharan, Backes, and
  McDaniel]{grosse_adversarial_2016}
Grosse, K., Papernot, N., Manoharan, P., Backes, M., and McDaniel, P.
\newblock Adversarial perturbations against deep neural networks for malware
  classification.
\newblock \emph{arXiv:1606.04435 [cs]}, June 2016.
\newblock arXiv: 1606.04435.

\bibitem[Ilyas et~al.(2017)Ilyas, Jalal, Asteri, Daskalakis, and
  Dimakis]{ilyas_robust_2017}
Ilyas, A., Jalal, A., Asteri, E., Daskalakis, C., and Dimakis, A.
\newblock The robust manifold defense: adversarial training using generative
  models.
\newblock \emph{arXiv:1712.09196 [cs, stat]}, December 2017.
\newblock arXiv: 1712.09196.

\bibitem[Ilyas et~al.(2018)Ilyas, Engstrom, Athalye, and
  Lin]{ilyas_black-box_2018}
Ilyas, A., Engstrom, L., Athalye, A., and Lin, J.
\newblock Black-box adversarial attacks with limited queries and information.
\newblock \emph{arXiv:1804.08598 [cs, stat]}, July 2018.
\newblock arXiv: 1804.08598.

\bibitem[Kang et~al.(2019)Kang, Sun, Brown, Hendrycks, and
  Steinhardt]{kang_transfer_2019}
Kang, D., Sun, Y., Brown, T., Hendrycks, D., and Steinhardt, J.
\newblock Transfer of adversarial robustness between perturbation types.
\newblock \emph{arXiv:1905.01034 [cs, stat]}, May 2019.
\newblock arXiv: 1905.01034.

\bibitem[Khormali et~al.(2019)Khormali, Abusnaina, Chen, Nyang, and
  Mohaisen]{khormali_copycat:_2019}
Khormali, A., Abusnaina, A., Chen, S., Nyang, D., and Mohaisen, A.
\newblock {COPYCAT}: practical adversarial attacks on visualization-based
  malware detection.
\newblock \emph{arXiv:1909.09735 [cs]}, September 2019.
\newblock arXiv: 1909.09735.

\bibitem[Ko et~al.(2019)Ko, Lyu, Weng, Daniel, Wong, and Lin]{ko_popqorn:_2019}
Ko, C., Lyu, Z., Weng, T., Daniel, L., Wong, N., and Lin, D.
\newblock {POPQORN}: quantifying robustness of recurrent neural networks.
\newblock \emph{arXiv:1905.07387 [cs, stat]}, May 2019.
\newblock arXiv: 1905.07387.

\bibitem[Kurakin et~al.(2016)Kurakin, Goodfellow, and
  Bengio]{kurakin_adversarial_2016}
Kurakin, A., Goodfellow, I., and Bengio, S.
\newblock Adversarial examples in the physical world.
\newblock \emph{arXiv:1607.02533 [cs.CV]}, July 2016.

\bibitem[Li et~al.(2019)Li, Lai, and Cui]{li_adversarial_2019-1}
Li, F., Lai, L., and Cui, S.
\newblock On the adversarial robustness of subspace learning.
\newblock \emph{arXiv:1908.06210 [cs, eess, stat]}, August 2019.
\newblock arXiv: 1908.06210.

\bibitem[Lin et~al.(2019)Lin, Song, He, Wang, and Hopcroft]{lin_nesterov_2019}
Lin, J., Song, C., He, K., Wang, L., and Hopcroft, J.
\newblock Nesterov accelerated gradient and scale invariance for improving
  transferability of adversarial examples.
\newblock \emph{arXiv:1908.06281 [cs, stat]}, August 2019.
\newblock arXiv: 1908.06281.

\bibitem[Liu et~al.(2016)Liu, Chen, Liu, and Song]{liu_delving_2016}
Liu, Y., Chen, X., Liu, C., and Song, D.
\newblock Delving into transferable adversarial examples and black-box attacks.
\newblock \emph{arXiv:1611.02770 [cs]}, November 2016.
\newblock arXiv: 1611.02770.

\bibitem[Madry et~al.(2017)Madry, Makelov, Schmidt, Tsipras, and
  Vladu]{madry_towards_2017}
Madry, A., Makelov, A., Schmidt, L., Tsipras, D., and Vladu, A.
\newblock Towards deep learning models resistant to adversarial attacks.
\newblock \emph{arXiv:1706.06083 [cs, stat]}, June 2017.
\newblock arXiv: 1706.06083.

\bibitem[Meng et~al.(2020)Meng, Su, O'Kane, and Jamshidi]{meng_ensembles_2020}
Meng, Y., Su, J., O'Kane, J., and Jamshidi, P.
\newblock Ensembles of many diverse weak defenses can be strong: defending deep
  neural networks against adversarial attacks.
\newblock \emph{arXiv:2001.00308 [cs, stat]}, January 2020.
\newblock arXiv: 2001.00308.

\bibitem[Meunier et~al.(2019)Meunier, Atif, and Teytaud]{meunier_yet_2019}
Meunier, L., Atif, J., and Teytaud, O.
\newblock Yet another but more efficient black-box adversarial attack: tiling
  and evolution strategies.
\newblock \emph{arXiv:1910.02244 [cs]}, October 2019.
\newblock arXiv: 1910.02244.

\bibitem[Moosavi-Dezfooli et~al.(2016)Moosavi-Dezfooli, Fawzi, and
  Frossard]{moosavi-dezfooli_deepfool:_2016}
Moosavi-Dezfooli, S., Fawzi, A., and Frossard, P.
\newblock {DeepFool}: a simple and accurate method to fool deep neural
  networks.
\newblock pp.\  2574--2582, 2016.

\bibitem[Narodytska \& Kasiviswanathan(2016)Narodytska and
  Kasiviswanathan]{narodytska_simple_2016}
Narodytska, N. and Kasiviswanathan, S.
\newblock Simple black-box adversarial perturbations for deep networks.
\newblock \emph{arXiv:1612.06299 [cs, stat]}, December 2016.
\newblock arXiv: 1612.06299.

\bibitem[Papernot et~al.(2017)Papernot, McDaniel, Goodfellow, Jha, Celik, and
  Swami]{papernot_practical_2017}
Papernot, N., McDaniel, P., Goodfellow, I., Jha, S., Celik, Z., and Swami, A.
\newblock Practical black-box attacks against machine learning.
\newblock In \emph{Proceedings of the 2017 {ACM} on {Asia} {Conference} on
  {Computer} and {Communications} {Security}}, {ASIA} {CCS} '17, pp.\
  506--519, New York, NY, USA, 2017. ACM.

\bibitem[Samangouei et~al.(2018)Samangouei, Kabkab, and
  Chellappa]{samangouei_defense-gan:_2018}
Samangouei, P., Kabkab, M., and Chellappa, R.
\newblock Defense-{GAN}: protecting classifiers against adversarial attacks
  using generative models.
\newblock \emph{arXiv:1805.06605 [cs, stat]}, May 2018.
\newblock arXiv: 1805.06605.

\bibitem[Song et~al.(2018)Song, Kim, Nowozin, Ermon, and
  Kushman]{song_pixeldefend:_2018}
Song, Y., Kim, T., Nowozin, S., Ermon, S., and Kushman, N.
\newblock {PixelDefend}: leveraging generative models to understand and defend
  against adversarial examples.
\newblock \emph{arXiv:1710.10766 [cs]}, May 2018.
\newblock arXiv: 1710.10766.

\bibitem[Szegedy et~al.(2013)Szegedy, Zaremba, Sutskever, Bruna, Erhan,
  Goodfellow, and Fergus]{szegedy_intriguing_2013}
Szegedy, C., Zaremba, W., Sutskever, I., Bruna, J., Erhan, D., Goodfellow, I.,
  and Fergus, R.
\newblock Intriguing properties of neural networks.
\newblock \emph{arXiv:1312.6199 [cs]}, December 2013.
\newblock arXiv: 1312.6199.

\bibitem[Tramer et~al.(2017)Tramer, Papernot, Goodfellow, Boneh, and
  McDaniel]{tramer_space_2017}
Tramer, F., Papernot, N., Goodfellow, I., Boneh, D., and McDaniel, P.
\newblock The space of transferable adversarial examples.
\newblock \emph{arXiv:1704.03453 [cs, stat]}, April 2017.
\newblock arXiv: 1704.03453.

\bibitem[Tu et~al.(2019)Tu, Ting, Chen, Liu, Zhang, Yi, Hsieh, and
  Cheng]{tu_autozoom_2019}
Tu, C., Ting, P., Chen, P., Liu, ., Zhang, H., Yi, J., Hsieh, C., and Cheng, S.
\newblock {AutoZOOM}: autoencoder-based zeroth order optimization method for
  attacking black-box neural networks.
\newblock \emph{Proceedings of the AAAI Conference on Artificial Intelligence},
  33\penalty0 (01):\penalty0 742--749, July 2019.
\newblock ISSN 2374-3468.
\newblock \doi{10.1609/aaai.v33i01.3301742}.

\bibitem[Wang et~al.(2020)Wang, Nepal, Rudolph, Grobler, Chen, and
  Chen]{wang_backdoor_2020}
Wang, S., Nepal, S., Rudolph, C., Grobler, M., Chen, S., and Chen, T.
\newblock Backdoor attacks against transfer learning with pre-trained deep
  learning models.
\newblock \emph{arXiv:2001.03274 [cs]}, January 2020.
\newblock arXiv: 2001.03274.

\bibitem[Xie et~al.(2018)Xie, Wu, van~der Maaten, Yuille, and
  He]{xie_feature_2018}
Xie, C., Wu, Y., van~der Maaten, L., Yuille, A., and He, K.
\newblock Feature denoising for improving adversarial robustness.
\newblock \emph{arXiv:1812.03411 [cs.CV]}, December 2018.

\bibitem[Yan et~al.(2019)Yan, Guo, and Zhang]{yan_subspace_2019}
Yan, Z., Guo, Y., and Zhang, C.
\newblock Subspace attack: exploiting promising subspaces for query-efficient
  black-box attacks.
\newblock \emph{arXiv:1906.04392 [cs]}, June 2019.
\newblock arXiv: 1906.04392.

\bibitem[Yi et~al.(2019)Yi, Xie, Zhou, Wu, Xu, and Mudumbai]{yi_trust_2019}
Yi, J., Xie, H., Zhou, L., Wu, X., Xu, W., and Mudumbai, R.
\newblock Trust but verify: an information-theoretic explanation for the
  adversarial fragility of machine learning systems, and a general defense
  against adversarial attacks.
\newblock \emph{arXiv:1905.11381 [cs, stat]}, May 2019.
\newblock arXiv: 1905.11381.

\bibitem[Yin et~al.(2019)Yin, Kolouri, and Rohde]{yin_divide-and-conquer_2019}
Yin, X., Kolouri, S., and Rohde, G.
\newblock Divide-and-conquer adversarial detection.
\newblock \emph{arXiv:1905.11475 [cs, stat]}, May 2019.
\newblock arXiv: 1905.11475.

\bibitem[Zhang \& Liang(2019)Zhang and Liang]{zhang_defending_2019}
Zhang, Y. and Liang, P.
\newblock Defending against whitebox adversarial attacks via randomized
  discretization.
\newblock \emph{arXiv:1903.10586 [cs, stat]}, March 2019.
\newblock arXiv: 1903.10586.

\bibitem[Zhao et~al.(2019)Zhao, Shumailov, Cui, Gao, Mullins, and
  Anderson]{zhao_blackbox_2019}
Zhao, Y., Shumailov, I., Cui, H., Gao, X., Mullins, R., and Anderson, R.
\newblock Blackbox attacks on reinforcement learning agents using approximated
  temporal information.
\newblock \emph{arXiv:1909.02918 [cs, stat]}, September 2019.
\newblock arXiv: 1909.02918.

\bibitem[Zheng et~al.(2019)Zheng, Zhang, Gu, Lee, and
  Prakash]{zheng_efficient_2019}
Zheng, H., Zhang, Z., Gu, J., Lee, H., and Prakash, A.
\newblock Efficient adversarial training with transferable adversarial
  examples.
\newblock \emph{arXiv:1912.11969 [cs]}, December 2019.
\newblock arXiv: 1912.11969.

\bibitem[Zhou et~al.(2019)Zhou, Guan, Bhat, and Hsu]{zhou_fake_2019}
Zhou, Z., Guan, H., Bhat, M., and Hsu, J.
\newblock Fake news detection via {NLP} is vulnerable to adversarial attacks.
\newblock \emph{arXiv:1901.09657 [cs]}, January 2019.
\newblock arXiv: 1901.09657.

\bibitem[Zügner \& Günnemann(2019)Zügner and
  Günnemann]{zugner_adversarial_2019}
Zügner, D. and Günnemann, S.
\newblock Adversarial attacks on graph neural networks via meta {Learning}.
\newblock \emph{arXiv:1902.08412 [cs, stat]}, February 2019.
\newblock arXiv: 1902.08412.

\end{thebibliography}

\end{document}